\documentclass[runningheads]{llncs}
\usepackage{graphicx}
\usepackage{amsmath,amssymb} 
\usepackage{color}
\usepackage[pagebackref=true,breaklinks=true,colorlinks,bookmarks=false]{hyperref}
\usepackage[width=122mm,left=12mm,paperwidth=146mm,height=193mm,top=12mm,paperheight=217mm]{geometry}
\usepackage{diagbox}
\usepackage{multirow}
\begin{document}
\title{Towards a Better Match in Siamese Network Based Visual Object Tracker} 

\titlerunning{Towards a Better Match in Siamese Network Based Visual Object Tracker}
%
\author{Anfeng He\thanks{This work is carried out while Anfeng He is an intern in MSRA.}\inst{1} \and
        Chong Luo\inst{2} \and
        Xinmei Tian\inst{1} \and
        Wenjun Zeng\inst{2}
       }
%
\authorrunning{Anfeng He, Chong Luo, Xinmei Tian and Wenjun Zeng}
%

\institute{CAS Key Laboratory of Technology in Geo-Spatial Information Processing and Application System, University of Science and Technology of China,\\
Hefei, Anhui, China\\
\email{heanfeng@mail.ustc.edu.cn}, \email{xinmei@ustc.edu.cn}\\
\and
Microsoft Research, Beijing, China\\
\email{\{cluo, wezeng\}@microsoft.com}}
\maketitle              
\begin{abstract}
Recently, Siamese network based trackers have received tremendous interest for their fast tracking speed and high performance. Despite the great success, this tracking framework still suffers from several limitations. First, it cannot properly handle large object rotation. Second, tracking gets easily distracted when the background contains salient objects. In this paper, we propose two simple yet effective mechanisms, namely angle estimation and spatial masking, to address these issues. The objective is to extract more representative features so that a better match can be obtained between the same object from different frames. The resulting tracker, named Siam-BM, not only significantly improves the tracking performance, but more importantly maintains the realtime capability. Evaluations on the VOT2017 dataset show that Siam-BM achieves an EAO of 0.335, which makes it the best-performing realtime tracker to date.  

\end{abstract}
    \section{Introduction}
    
    Generic visual object tracking is a challenging and fundamental task in the area of computer vision and artificial intelligence. A tracker is initialized with only the bounding box of an unknown target in the first frame. The task of the tracker is to predict the bounding boxes of the target in the following frames. There are numerous applications of object tracking, such as augmented reality, surveillance and autonomous systems. However, robust and precise tracking is still an open problem. 
    
In the past a few years, with the penetration of deep convolutional neural networks (DCNN) in various vision problems, there emerge a large number of DCNN-based trackers \cite{SASiam,SiamFC,DeepSRDCF,MDNET,ECO,SINT,SiamDCF,HCF,BranchOut,CSRDCF,GOTURN,CCOT}, among which the siamese network based trackers have received great attention. The pioneering work in this category is the SiamFC tracker \cite{SiamFC}. The basic idea is to use the same DCNN to extract features from the target image patch and the search region, and to generate a response map by correlating the two feature maps. The position with the highest response indicates the position of the target object in the search region. The DCNN is pre-trained and remains unchanged during testing time. This allows SiamFC to achieve high tracking performance in realtime. Follow-up work of SiamFC includes SA-Siam, SiamRPN, RASNet, EAST, DSiam, CFNET and SiamDCF \cite{SASiam,SiamRPN,RASNet,EAST,DSiam,CFNET,SiamDCF}. 
    \begin{figure}[t!]
        \begin{center}
        \includegraphics[width=0.9\columnwidth]{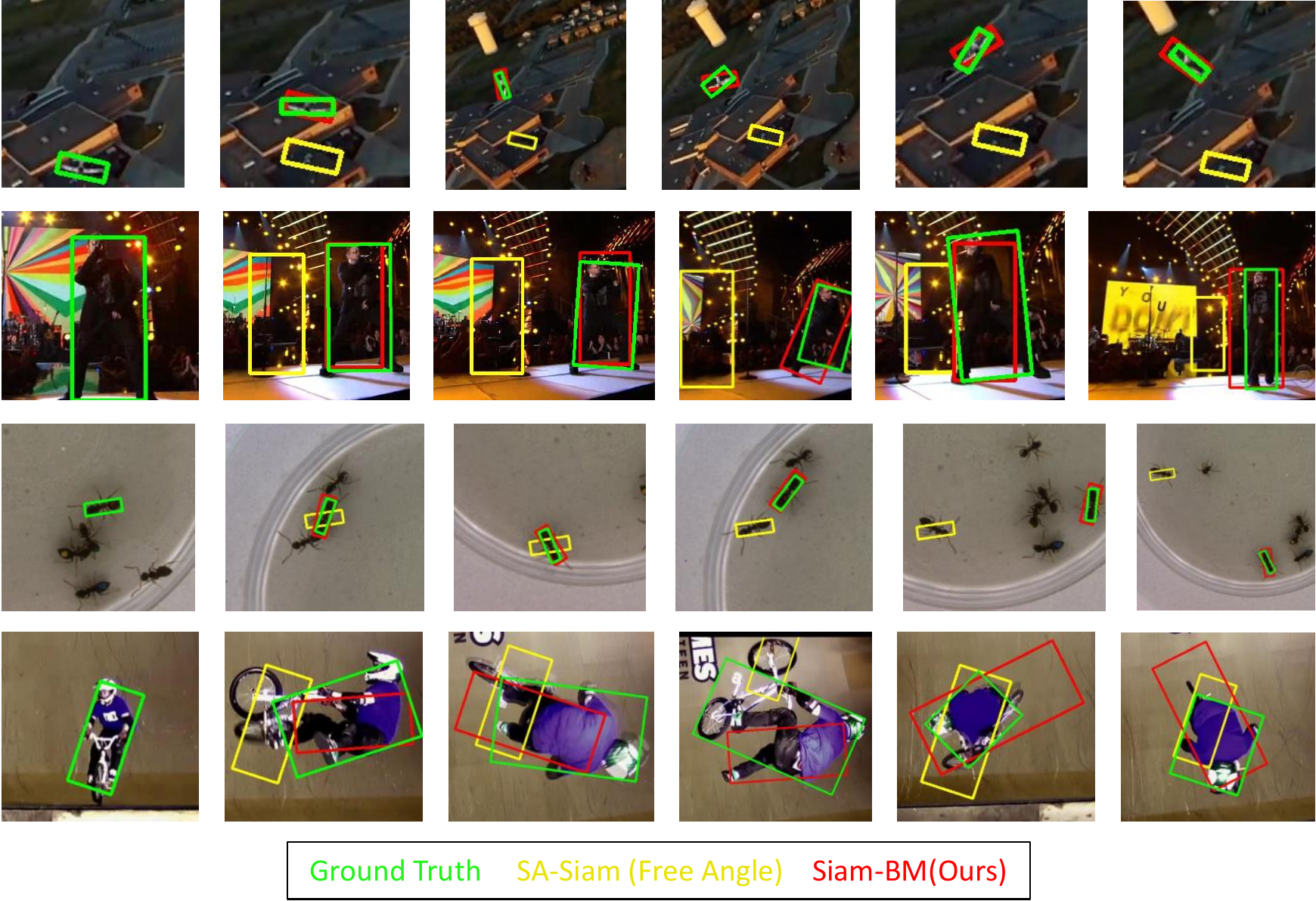}
        \end{center}
        \caption{Comparison with our tracker and baseline tracker. Best view in color.}
        \label{fig:comp_baseline_ours_gallery}
    \end{figure}
    
    Despite the great success of siamese network-based trackers, there are still some limitations in this framework. 
    First, as previous research \cite{ORN,TICNN,STN} pointed out, the CNN features are not invariant to large image transformations such as scaling and rotation. Therefore, SiamFC does not perform well when the object has large scale change or in-plane rotation. This problem is exaggerated when the tracked object is non-square, because there is no mechanism in the SiamFC framework that can adjust the orientation or the aspect ratio of the tracked object bounding box.
    Second, it is hard to determine the spatial region from which DNN features should be extracted to represent the target object. Generally speaking, including a certain range of the surrounding context is helpful to tracking, but too many of them could be unprofitable especially when the background contains distracting objects. Recently, Wang et al. \cite{RASNet} also observed this problem and they propose to train a feature mask to highlight the features of the target object.

    In this paper, we revisit the SiamFC tracking framework and propose two simple yet effective mechanisms to address the above two issues. The computational overhead of these two mechanisms is kept low, such that the resulting tracker, named Siam-BM, can still run in real-time on GPU. 
    
    First, our tracker not only predicts the location and the scale of the target object, but also predicts the angle of the target object. This is simply achieved by enumerating several angle options and computing DCNN features for each option. However, in order to maintain the high speed of the tracker, it is necessary to trim the explosive number of (scale, angle) combinations without tampering the tracking performance. 
    Second, we propose to selectively apply a spatial mask to CNN feature maps when the possibility of distracting background objects is high. We apply such a mask when the aspect ratio of the target bounding box is far apart from 1:1. This simple mechanism not only saves the efforts to train an object-specific mask, but allows the feature map to include a certain amount of information of the background, which is in general helpful to tracking. 
    Last, we also adopt a simple template updating mechanism to cope with the gradual appearance change of the target object. All these mechanisms are toward the same goal to achieve a better match between the same object from different frames. Therefore, the resulting tracker is named Siam-BM.
    
    We carry out extensive experiments for the proposed Siam-BM tracker, over both the OTB and the VOT benchmarks. Results show that Siam-BM achieves an EAO of 0.335 at the speed of 32 fps on VOT-2017 dataset. It is the best-performing realtime tracker in literature.  

    The rest of the paper is organized as follows. We review related work in Section \ref{sec:related}. In Section \ref{sec:Siam-BM}, we revisit the SiamFC tracking framework and explain the proposed two mechanisms in details. Section \ref{sec:exp} provides implementation details of Siam-BM and presents the experimental results. We finally conclude with some discussions in Section \ref{sec:conclude}. 
    
\section{Related Work}\label{sec:related}
    Visual object tracking is an important computer vision problem. It can be modeled as a similarity matching problem. 
In recent years, with the widespread use of deep neural networks, there emerge a bunch of Siamese network based trackers, which performs similarity matching based on extracted DCNN features. The pioneering work in this category is the fully convolutional Siamese network (SiamFC) \cite{SiamFC}. SiamFC extract DCNN features from the target patch and the search region using AlexNet. Then, a response map is generated by correlating the two feature maps. The object is tracked to the location where the highest response is obtained. A notable advantage of this method is that it needs no or little online training. Thus, real-time tracking can be easily achieved.  
    
There are a large number of follow-up work \cite{RFL,CFNET,EAST,SINT,DSiam,SiamRPN,SiamDCF,SASiam,HP,RASNet} of SiamFC. EAST \cite{EAST} attempts to speed up the tracker by early stopping the feature extractor if low-level features are sufficient to track the target. CFNet \cite{CFNET} introduces correlation filters for low level CNNs features to speed up tracking without accuracy drop. 
SINT \cite{SINT} incorporates the optical flow information and achieves better performance. However, since computing optical flow is computationally expensive, SINT only operates at 4 frames per second (fps). DSiam \cite{DSiam} attempts to online update the embeddings of tracked target. Significantly better performance is achieved without much speed drop. HP \cite{HP} tries to tune hyperparameters for each sequence in SiamFC \cite{SiamFC} by optimizing it with continuous Q-Learning. 
RASNet \cite{RASNet} introduces three kinds of attention mechanisms for SiamFC \cite{SiamFC} tracker. The authors share the same vision with us to look for more precise feature representation for the tracked object. 
SiamRPN \cite{SiamRPN} includes a region proposal subnetwork to estimate the aspect ratio of the target object. This network will generate a more compact bounding box when the target shape changes. 
SA-Siam \cite{SASiam} utilizes complementary appearance and semantic features to represent the tracked object. A channel-wise attention mechanism is used for semantic feature selection. SA-Siam achieves a large performance gain at a small computational overhead. 

Apparently we are not the first who concerns transformation estimation in visual object tracking. In correlation filter based trackers, DSST \cite{DSST} and SAMF \cite{SAMF} are early work that estimates the scale change of the tracked object. DSST \cite{DSST} does so by learning separate discriminative correlation filters for translation and scale estimation. SAMF \cite{SAMF} uses a scale pyramid to search corresponding target scale. Recently, RAJSSC \cite{RAJSSC} proposes to perform both scale and angle estimation in a unified correlation tracking framework by using the Log-Polar transformation. In SiamFC-based trackers, while the scale estimation has been considered in the original SiamFC tracker, angle estimation has not been considered before. 

There are also a couple of previous research efforts to suppress the background noise. SRDCF \cite{SRDCF} and DeepSRDCF \cite{DeepSRDCF} reduce background noise by introducing the spatial regularization term in loss function during the online training of correlation filters. RASNet \cite{RASNet} and SA-Siam \cite{SASiam} are two SiamFC-based trackers. They adopt spatial attention or channel-wise attention in the feature extraction network. They both need careful training of the attention blocks. 
    
\section{Siam-BM Tracker} \label{sec:Siam-BM}

Our tracker Siam-BM is built upon the recent SA-Siam tracker \cite{SASiam}, which is in turn built upon the SiamFC tracker \cite{SiamFC}. The main difference between SA-Siam and SiamFC trackers is that the former extracts semantic features in addition to appearance features for similarity matching. In this section, we will first revisit the SiamFC tracking framework and then present the two proposed mechanisms in Siam-BM towards a better matching of object features.

\subsection{An Overview of the SiamFC Tracking Framework}

\begin{figure}[th!]
    \begin{center}
    \includegraphics[width=0.7\columnwidth]{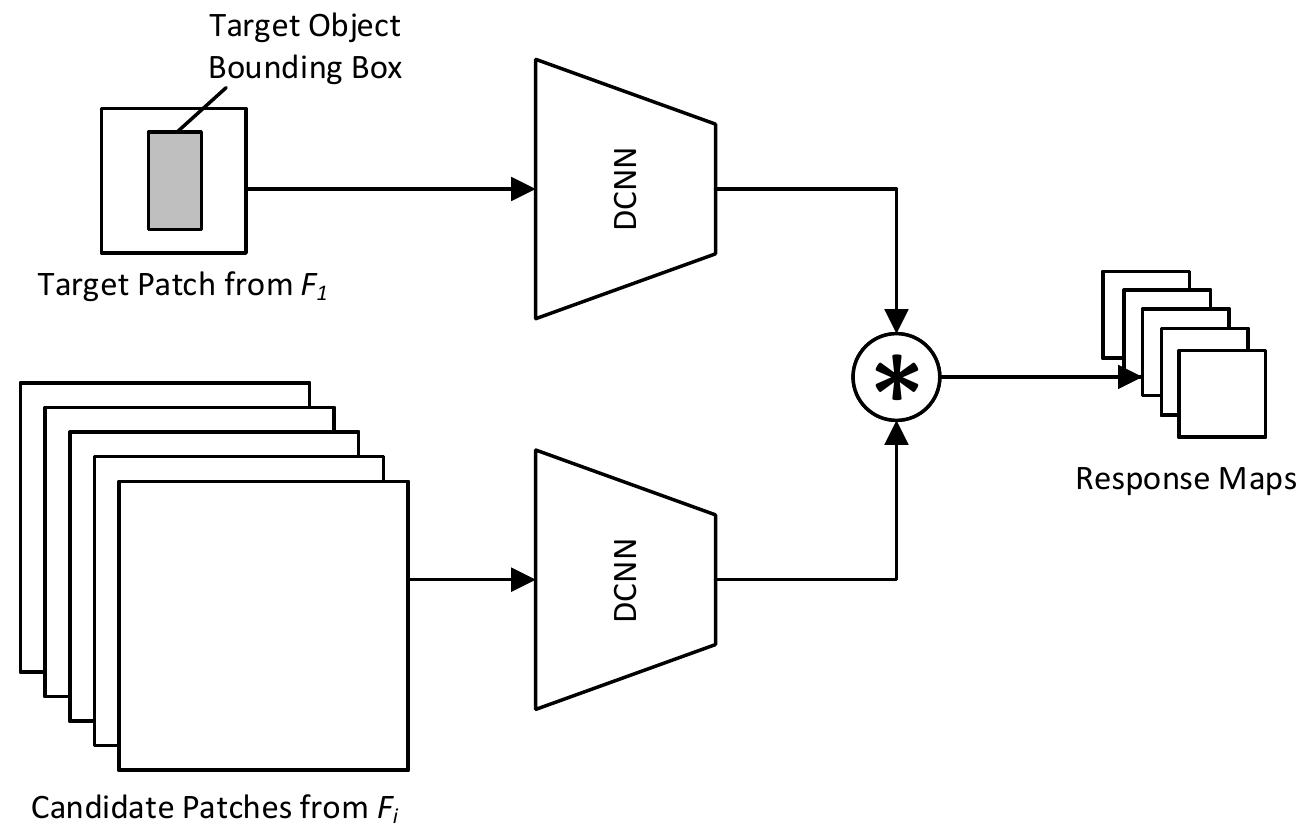}
    \end{center}
    \caption{The SiamFC Tracking Framework}
    \label{fig:siamFC}
\end{figure}

Fig.\ref{fig:siamFC} shows the basic operations in the SiamFC tracking framework. The input of the tracker is the target object bounding box $B_0$ in the first frame $F_1$. A bounding box can be described by a four-tuple $(x, y, w, h)$, where $(x, y)$ is the center coordinates and $w$, $h$ are the width and the height, respectively. SiamFC crops the target patch $T$ from the first frame, which is a square region covering $B_0$ and a certain amount of surrounding context. When the tracker comes to the $i^{th}$ frame, several candidate patches $\{C_1, C_2, ... C_M\}$ are drawn, all of which are centered at the tracked location of the previous frame, but differ in scales. In the original SiamFC \cite{SiamFC} work, $M$ is set to 3 or 5 to deal with 3 or 5 different scales. 

Both the target patch and the candidate patches go through the same DCNN network, which is fixed during testing time. The process of extracting DCNN features can be described by a function $\phi(\cdot)$. Then, $\phi(T)$ is correlated with $\phi(C_1)$ through $\phi(C_M)$ and $M$ response maps $\{R_1, R_2, ... R_M \}$ are computed. The position with the highest value in the response maps is determined by:
\begin{equation}
(x_i, y_i, m_i) = \mathop{\arg \max}_{x, y, m} R_m , \quad (m=1...M),
\end{equation}
where $x_i, y_i$ are the coordinates of the highest-value position and $m$ is the index of the response map from which the highest value is found. Then, the tracking result is given by $B_i = (x_i, y_i, s_{m_i}\cdot w, s_{m_i}\cdot h)$, where $s_{m_i}$ is the scale factor of the $m_i^{th}$ candidate patch. 

In this process, SiamFC tracker only determines the center position and the scale of the target object, but keeps the orientation and aspect ratio unchanged. This becomes a severe limitation of SiamFC tracker. 

\subsection{Angle Estimation}

As previous research \cite{ORN,STN,TICNN} has pointed out, DCNN features are not invariant to large image transformations, such as scaling and rotation. While scaling has been handled in the original SiamFC tracker, the rotation of the target object is not considered. Ideally, the change of object angle, or object rotation, can be similarly addressed as object scaling. Specifically, one could enumerate several possible angle changes and increase the number of candidate patches for similarity matching. However, with $M$ scale choices and $N$ angle choices, the number of candidate patches becomes $M \times N$. It is quite clear that the tracker speed is inversely proportional to the number of candidate patches. Using contemporary GPU hardware, a SiamFC tracker becomes non-realtime even when $M=N=3$.

Knowing the importance of realtime tracking, we intend to find a mechanism to reduce the number of candidate patches without tampering the performance of the tracker. The solution turns out to be a simple one: the proposed Siam-BM tracker adjusts the properties (scale or angle) of the tracked object only one at a time. In other words, Siam-BM can adjust both scale and angle in two frames, if necessary. As such, the number of candidate patches is reduced from $M \times N$ to $M+N-1$. In our implementation, $M=N=3$, so only 5 candidate patches are involved in each tracking process. 

\begin{figure}[th]
    \begin{center}
    \includegraphics[width=0.95\columnwidth]{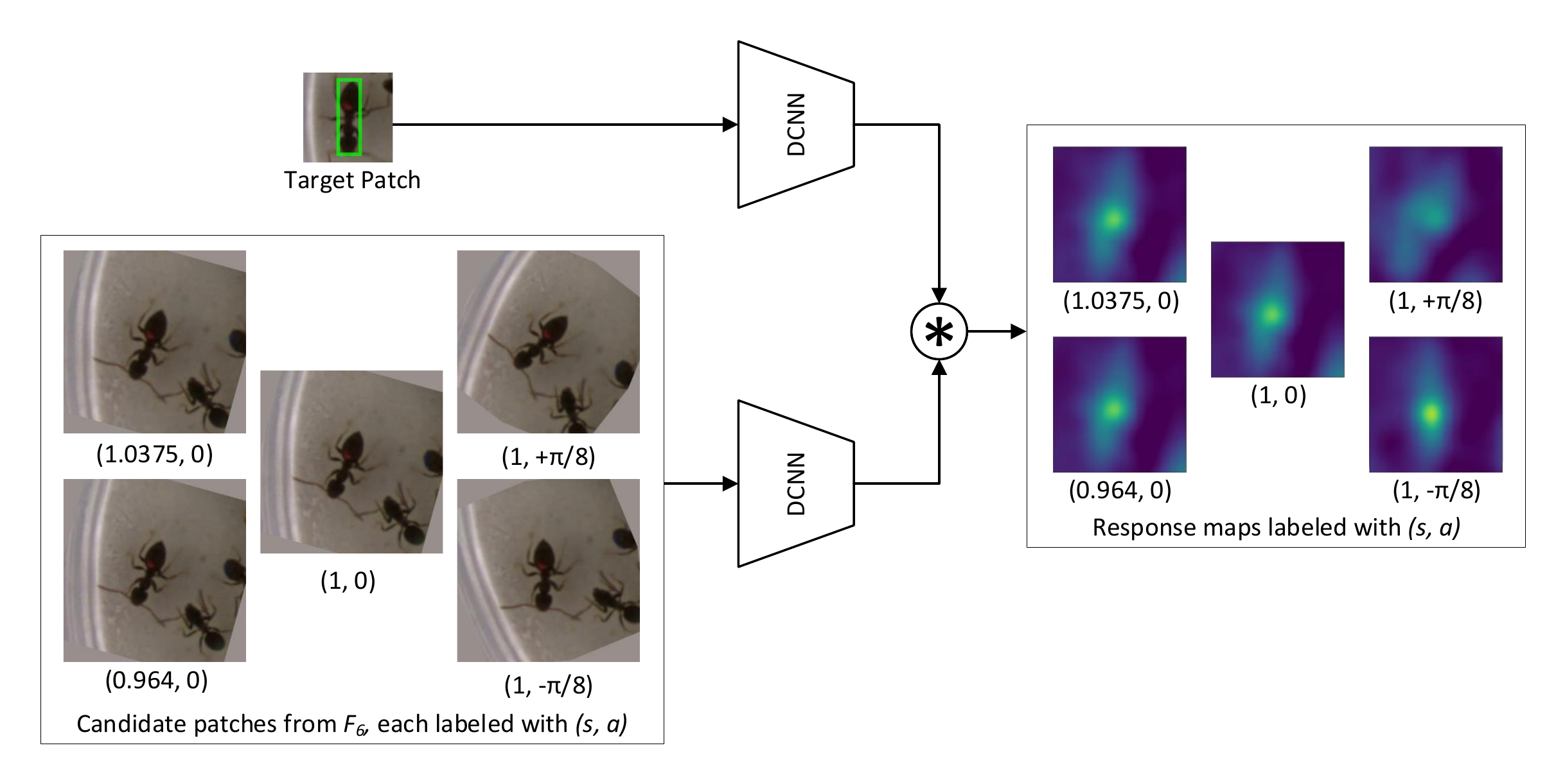} 
    \end{center}
    \caption{Illustrating the scale and angle estimation in Siam-BM.}
    \label{fig:angle_est}
\end{figure}

Mathematically, each candidate patch is associated with an $(s, a)$ pair, where $s$ is the scaling factor and $a$ is the rotation angle. It is forced that $s=1$ (no scale change) when $a \neq 0$ (angle change), and $a=0$ when $s \neq 1$. Similarly, the tracked object is determined by: 
\begin{equation}
(x_i, y_i, k_i) = \mathop{\arg \max}_{x, y, k} R_k , \quad (k=1, 2, ...K),
\end{equation}
where $K=M+N-1$ is the number of candidate patches. $(x_i, y_i)$ gives the center location of the tracked object and $k_i$ is associated with an $(s,a)$ pair, giving an estimation of the scale and angle changes. Both types of changes are accumulated during the tracking process. 

Fig.\ref{fig:angle_est} illustrates the scale and angle estimation in the proposed Siam-BM tracker. In the figure, each candidate patch and each response map are labeled with the corresponding $(s,a)$ pair. We can find that, when the target object has the same orientation in the target patch as in the candidate patch, the response is dramatically increased. In this example, the highest response in the map with $(1, -\pi/8)$ is significantly higher than the top values in other maps. 

\subsection{Spatial Mask}
    
\begin{figure}[th!]
    \begin{center}
    \includegraphics[width=0.95\columnwidth]{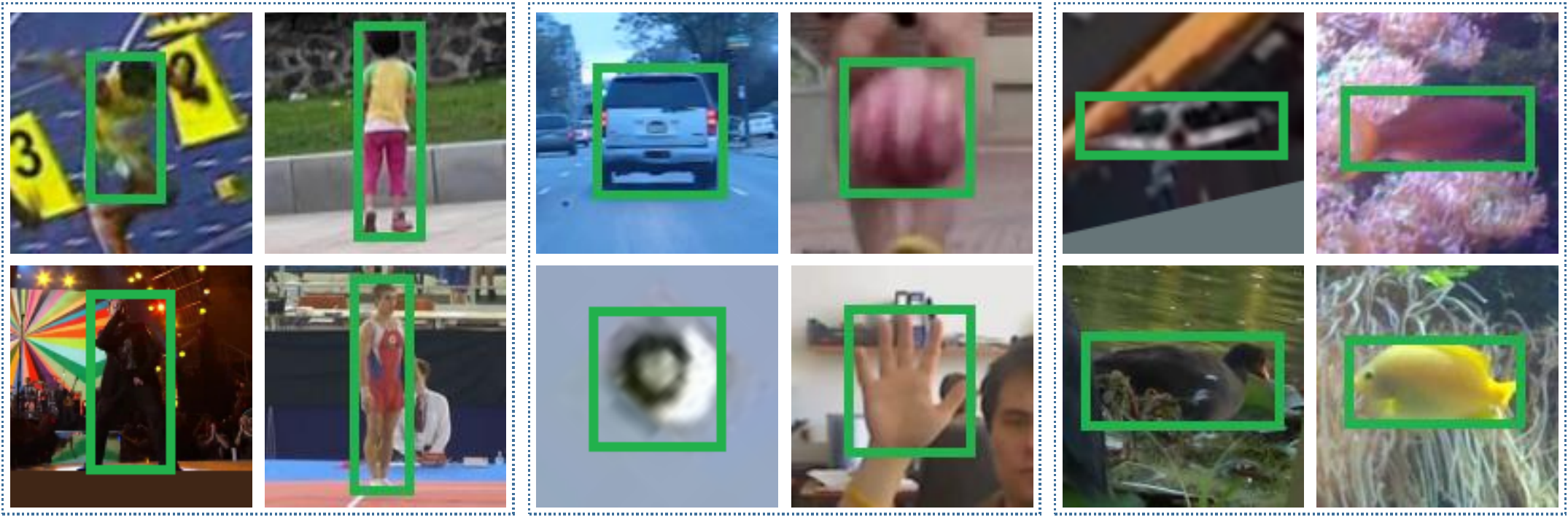}
    \end{center}
    \caption{Some examples of target patches containing objects with different aspect ratios. Target patches tend to include salient background objects when the object aspect ratio is far apart from 1:1. }
    \label{fig:example_ar}
\end{figure}

Context information is helpful during tracking. However, including too much context information could be distracting when the background has salient objects or prominent features. In the SiamFC framework, the target patch is always a square whose size is determined only by the area of the target object. Fig.\ref{fig:example_ar} shows some examples of target patches containing objects with different aspect ratios. It can be observed that, when the target object is a square, the background is made up of narrow stripes surrounding the target object, so the chance of having an integral salient object in it is small. But when the aspect ratio of the target object is far apart from 1 (vertical or horizontal), it is more likely to have salient objects in the background area. 

We propose to selectively apply spatial mask to the target feature map. In particular, when the aspect ratio of the target object exceeds a predefined threshold $th_r$, a corresponding mask is applied. 
We have mentioned that the proposed Siam-BM tracker is built upon a recent tracker named SA-Siam \cite{SASiam}. In SA-Siam, there is an attention module which serves a similar purpose. However, we find that the spatial mask performs better and is more stable than the channel-wise attention scheme. Therefore, we replace the channel attention model in SA-Siam with spatial masking in Siam-BM.

\subsection{The Siam-BM Tracker}    

Siam-BM is built upon SA-Siam \cite{SASiam}, which contains a semantic branch and an appearance branch for feature extraction. The target patch has a size of $127 \times 127$ as in SiamFC, and the candidate patches have a size of $255 \times 255$. We set $M=N=3$, so that there are 5 candidate patches and their corresponding scale and angle settings are $(s, a) = (1.0375, 0), (0.964, 0), (1, 0), (1, \pi/8), (1, -\pi/8)$. Correspondingly, five response maps are generated after combining semantic and appearance branches. Similar to SiamFC and SA-Siam, normalization and cosine window are applied to each of the five response maps. An angle penalty of $0.975$ is applied when $a \neq 0$ and a scale penalty of $0.973$ is applied when $s \neq 1$. 

\begin{figure}[th!]
    \begin{center}
    \includegraphics[width=0.6\columnwidth]{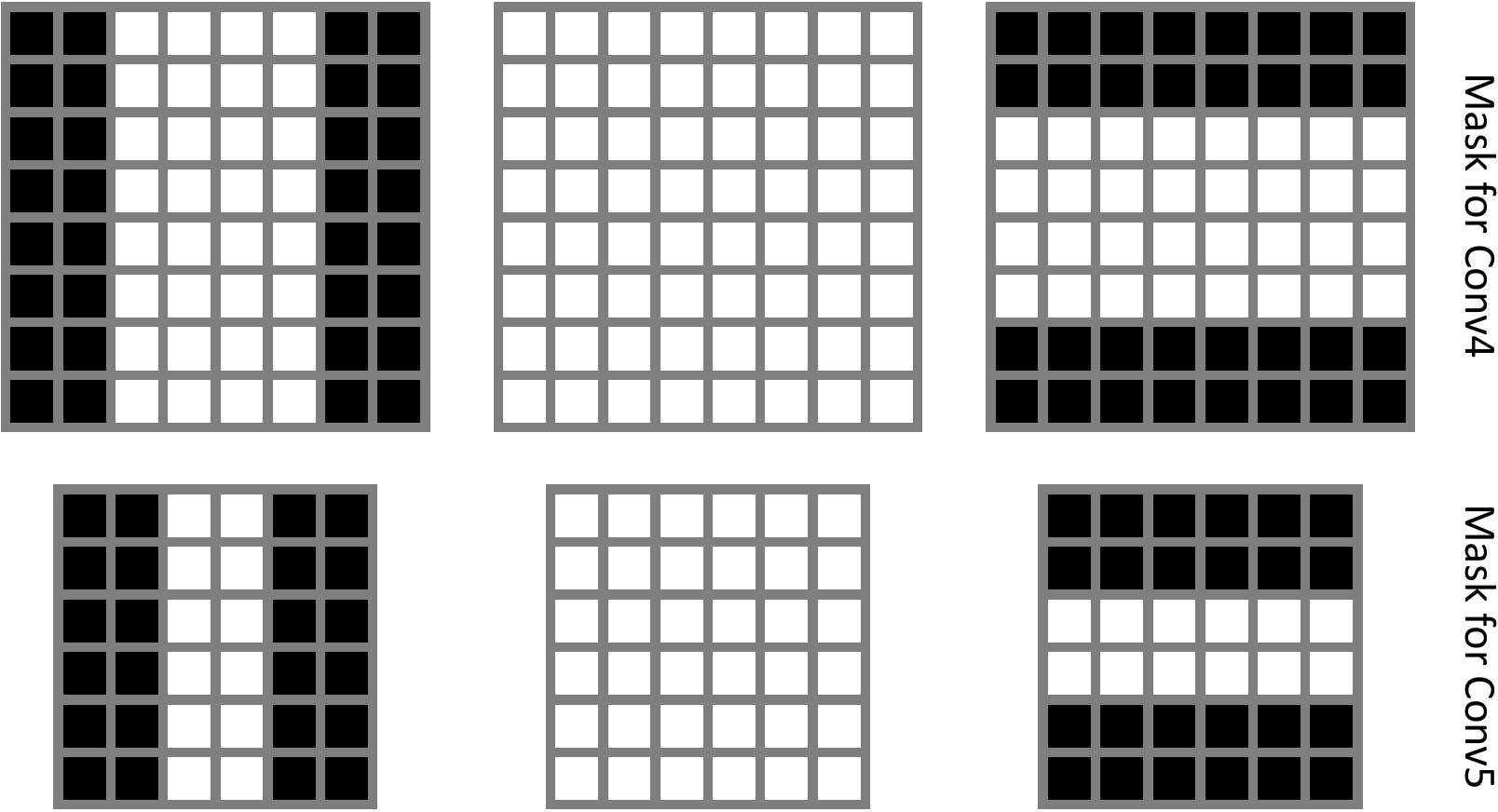}
    \end{center}
    \caption{Spatial feature mask when the aspect ratio of target object exceeds a predefined threshold. Left two masks: $h/w > th_r$; right two masks: $w/h > th_r$; middle two masks: $\max\{w/h, h/w\} < th_r$. }
    \label{fig:mask}
\end{figure}

Following SA-Siam, both conv4 and conv5 features are used, and the spatial resolutions are $8 \times 8$ and $6 \times 6$, respectively. Spatial mask is applied when the aspect ratio is greater than $th_r = 1.5$. Fig.\ref{fig:mask} shows the fixed design of spatial masks when $\max\{\frac{w}{h}, \frac{h}{w}\} > th_r$. The white grids indicate a coefficient of 1 and the black grids indicate a coefficient of 0. 

In addition, we perform template updating in Siam-BM. The template for frame $t$, denoted by $\phi(T_t)$ is defined as followings:
\begin{equation}
\phi(T_t) = \lambda_S \times \phi(T_1) + (1 - \lambda_S) \times \phi(T^u_t), 
\end{equation}
\begin{equation}
\phi(T^u_t) = (1 - \lambda_U) \times \phi(T^u_{t - 1}) + \lambda_U \times \hat{\phi}({T}_{t-1}).
\end{equation}
where $\hat{\phi}({T}_{t-1})$ is the feature of the tracked object in frame $t-1$. It can be cropped from the feature map of candidate regions of frame $t-1$. $\phi(T^u_t)$ is the moving average of feature maps with updating rate $\lambda_U$. $\lambda_S$ is the weight of the first frame. In our implementation, we set $\lambda_S = 0.5$, $\lambda_U = 0.006$. 

Note that the spatial mask is only applied to the semantic branch. This is because semantic responses are more sparse and centered than appearance responses, and it is less likely to exclude important semantic responses with spatial masks. The attention module in SA-Siam is removed.

\section{Experiments}
\label{sec:exp}
In this section, we evaluate the performance of Siam-BM tracker against state-of-the-art realtime trackers and carry out ablation studies to validate the contribution of angle estimation and spatial masking.

\subsection{Datasets and Evaluation Metrics}
    
\textbf{OTB: }The object tracking benchmarks (OTB) \cite{OTB13,OTB15} consist of two major datasets, namely OTB-2013 and OTB-100, which contain 51 and 100 sequences respectively. The two standard evaluation metrics on OTB are success rate and precision. For each frame, we compute the IoU (intersection over union) between the tracked and the groundtruth bounding boxes, as well as the distance of their center locations. A success plot can be obtained by evaluating the success rate at different IoU thresholds. Conventionally, the area-under-curve (AUC) of the success plot is reported. The precision plot can be acquired in a similar way, but usually the representative precision at the threshold of 20 pixels is reported. 
	    
\textbf{VOT:} We use the recent version of the VOT benchmark, denoted by VOT2017 \cite{VOT2017}. The VOT benchmarks evaluate a tracker by applying a reset-based methodology. Whenever a tracker has no overlap with the ground truth, the tracker will be re-initialized after five frames. Major evaluation metrics of VOT benchmarks are accuracy (A), robustness (R) and expected average overlap (EAO). A good tracker has high A and EAO scores but low R scores. 
    
In addition to the evaluation metrics, VOT differs from OTB in groundtruth labeling. In VOT, the groundtruth bounding boxes are not always upright. Therefore, we only evaluate the full version of Siam-BM on VOT. OTB is used to validate the effectiveness of spatial mask. 

\subsection{Training Siam-BM}
    
Similar to SA-Siam, the appearance network and the fuse module in semantic branch are trained using the ILSVRC-2015 video dataset (only color images are used). The semantic network uses the pretrained model for image classification on ILSVRC. Among a total of more than 4,000 sequences, there are around 1.3 million frames and about 2 million tracked objects with groundtruth bounding boxes. We strictly follow the separate training strategy in SA-Siam and the two branches are not combined until testing time. 

We implement our model in TensorFlow \cite{Tensorflow} 1.7.0 framework in Python 3.5.2 environment. Our experiments are performed on a PC with a Xeon E5-2690 2.60GHz CPU and a Tesla P100 GPU. 

\subsection{Ablation Analysis}

\textbf{Angle estimation:} We first evaluate whether angle estimation improves the performance on the VOT benchmark. Spatial masking is not added, so our method is denoted by Siam-BM (w/o mask). There are two baseline methods. In addition to vanilla SA-Siam, we implement a variation of SA-Siam, denoted by SA-Siam (free angle). Specifically, when the bounding box of the tracked object is not upright in the first frame, the reported tracking results are tilted by the same angle in all the subsequent frames. Table~\ref{table:poly_cmp} shows the EAO as well as accuracy and robustness of the three comparing schemes. Note that the performance of SA-Siam is slightly better than that reported in their original paper, which might due to some implementation differences. We can find that angle estimation significantly improves the tracker performance even when it is compared with the free angle version of SA-Siam.

\setlength{\tabcolsep}{4pt}
\begin{table}
\begin{center}
\caption{Comparison between Siam-BM (w/o mask) and two baseline trackers shows the effectiveness of angle estimation. }
\label{table:poly_cmp}
\begin{tabular}{|l|ccc|}
\hline
Trackers & EAO & Accuracy & Robustness\\
\hline
SA-Siam (vanilla) & 0.261 & 0.505 & 1.276 \\
SA-Siam (free angle) & 0.287 & 0.529& 1.234\\
\hline
Siam-BM (w/o mask)  & 0.301 & 0.544& 1.305\\
\hline
\end{tabular}
\end{center}
\end{table}
\setlength{\tabcolsep}{1.4pt}

\textbf{Spatial mask:} We use the OTB benchmark for this ablation study. Angle estimation is not added to the trackers evaluated in this part, therefore our method is denoted by Siam-BM (mask only). For all the 100 sequences in OTB benchmark, we compute the aspect ratio of the target object using $r = \max(\frac{h}{w}, \frac{w}{h})$, where $w$ and $h$ are the width and height of the groundtruth bounding box in the first frame. We set a threshold $th_r$, and if $r > th_r$, the object is called an \emph{elongated object}. Otherwise, we call the object a \emph{mediocre object}. At the testing stage, Siam-BM (mask only) applies spatial mask to elongated objects. At the training stage, we could either use the full feature map or the masked feature map for elongated objects. For mediocre objects, mask is not applied in training or testing. The comparison between different training and testing choices are included in Table~\ref{table:mask_train_test_cmp}.
Comparing (3)(4) with (5)(6) in the Table, we can conclude that applying spatial mask significantly improves the tracking performance for elongated objects. Comparison between (3) and (4) shows that training with spatial mask will further improve the performance for elongated objects, which agrees with the common practice to keep the consistency of training and testing. 
An interesting finding is obtained when we comparing (1) with (2). If we apply spatial mask to elongated objects during training, the Siamese network seems to be trained in a better shape and the tracking performance for mediocre objects is also improved even though no spatial mask is applied during testing time.     

\setlength{\tabcolsep}{8pt}
\begin{table}
\begin{center}
\caption{Comparison between training and testing choices with or without spatial mask. }
\label{table:mask_train_test_cmp}
\begin{tabular}{|c|c|c|c|c|}
\hline
\multirow{2}{*}{\diagbox{Training}{Testing}} & {mediocre object} & \multicolumn{2}{c|}{elongated object} \\
\cline{2-4}
&{no mask}  & {w/ mask}  & {w/o mask}\\
\hline
w/ mask & {0.681 (1)} & 0.654 (3)  & 0.581 (5)\\
\hline
w/o mask         & {0.665 (2)} & 0.644 (4) & 0.609 (6)\\
\hline
\end{tabular}
\end{center}
\end{table}
\setlength{\tabcolsep}{1.4pt}

\begin{figure}[th!]
    \begin{center}
    \includegraphics[width=0.45\columnwidth]{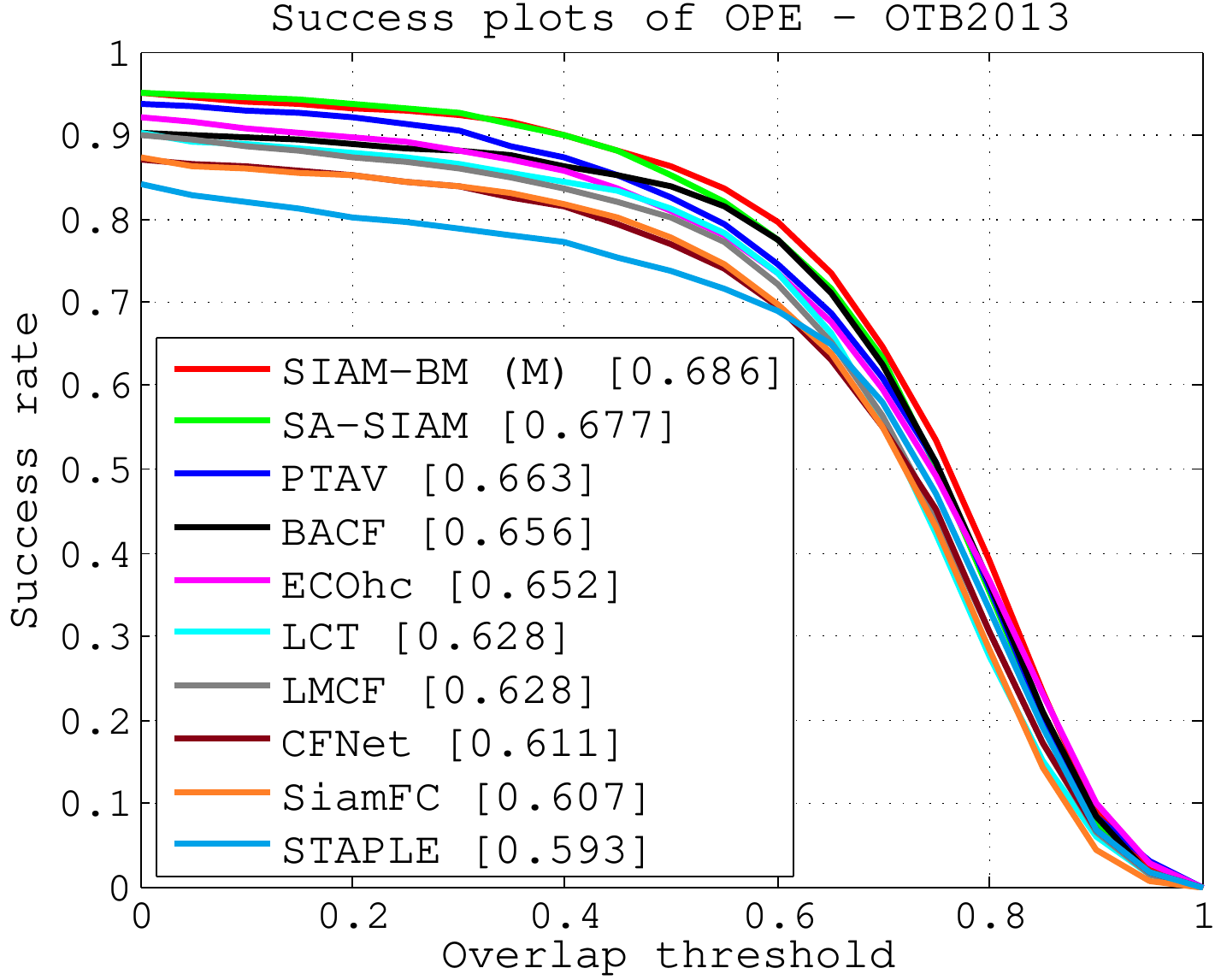}
    \includegraphics[width=0.45\columnwidth]{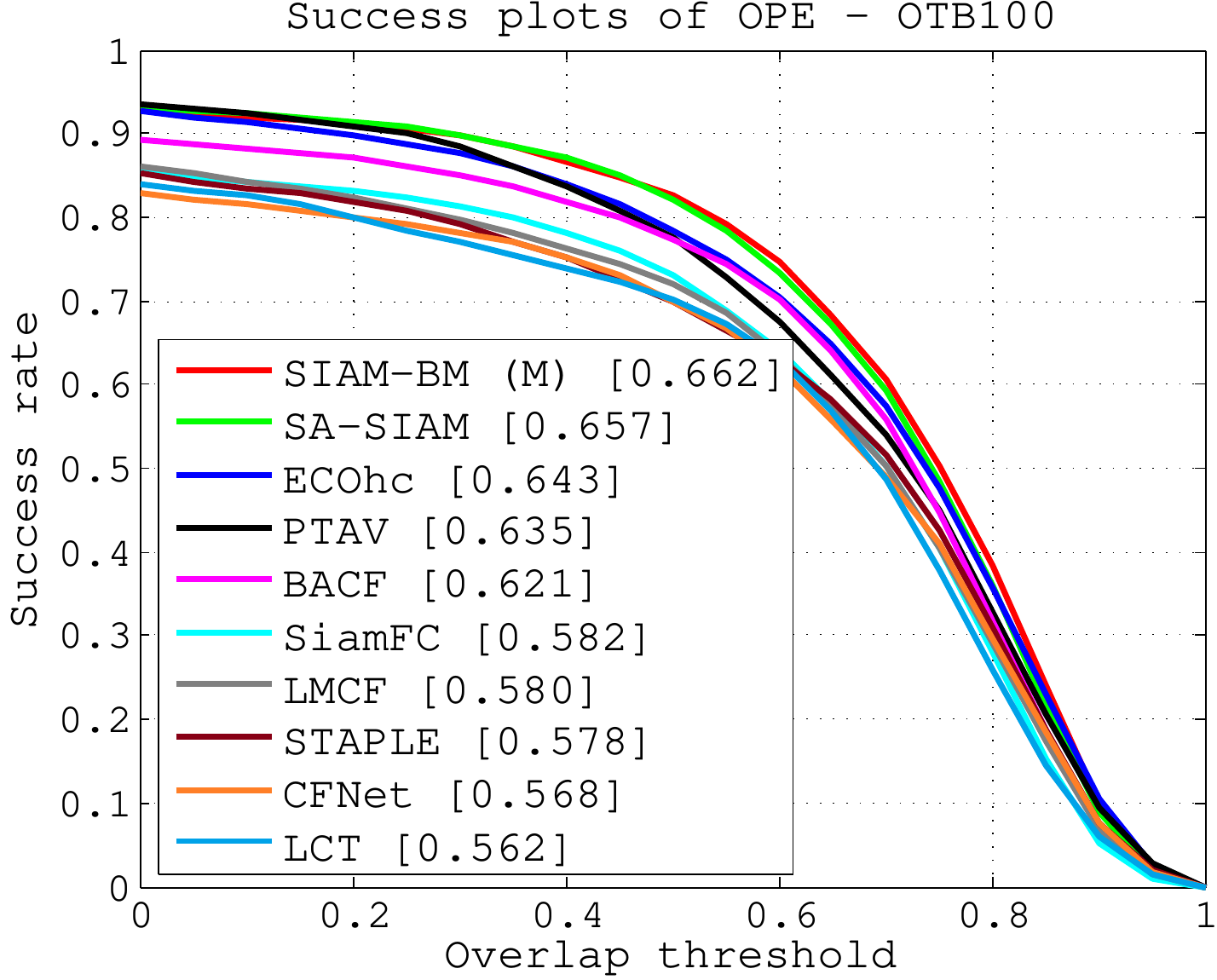}
    \end{center}
    \caption{Comparing Siam−BM (Mask only) with other high performance and real-time trackers}
    \label{fig:otb}
\end{figure}

We then compare the performance of Siam-BM (mask only) with the state-of-the-art realtime trackers on OTB-2013 and OTB-100, and the results are shown in Table~\ref{table:mask_cmp_stoa} and Fig.\ref{fig:otb}. The improvement of Siam-BM (mask only) with respect to SA-Siam demonstrates that the simple spatial masking mechanism is indeed effective.     

\begin{figure}[th!]
    \begin{center}
    \includegraphics[width=0.95\columnwidth]{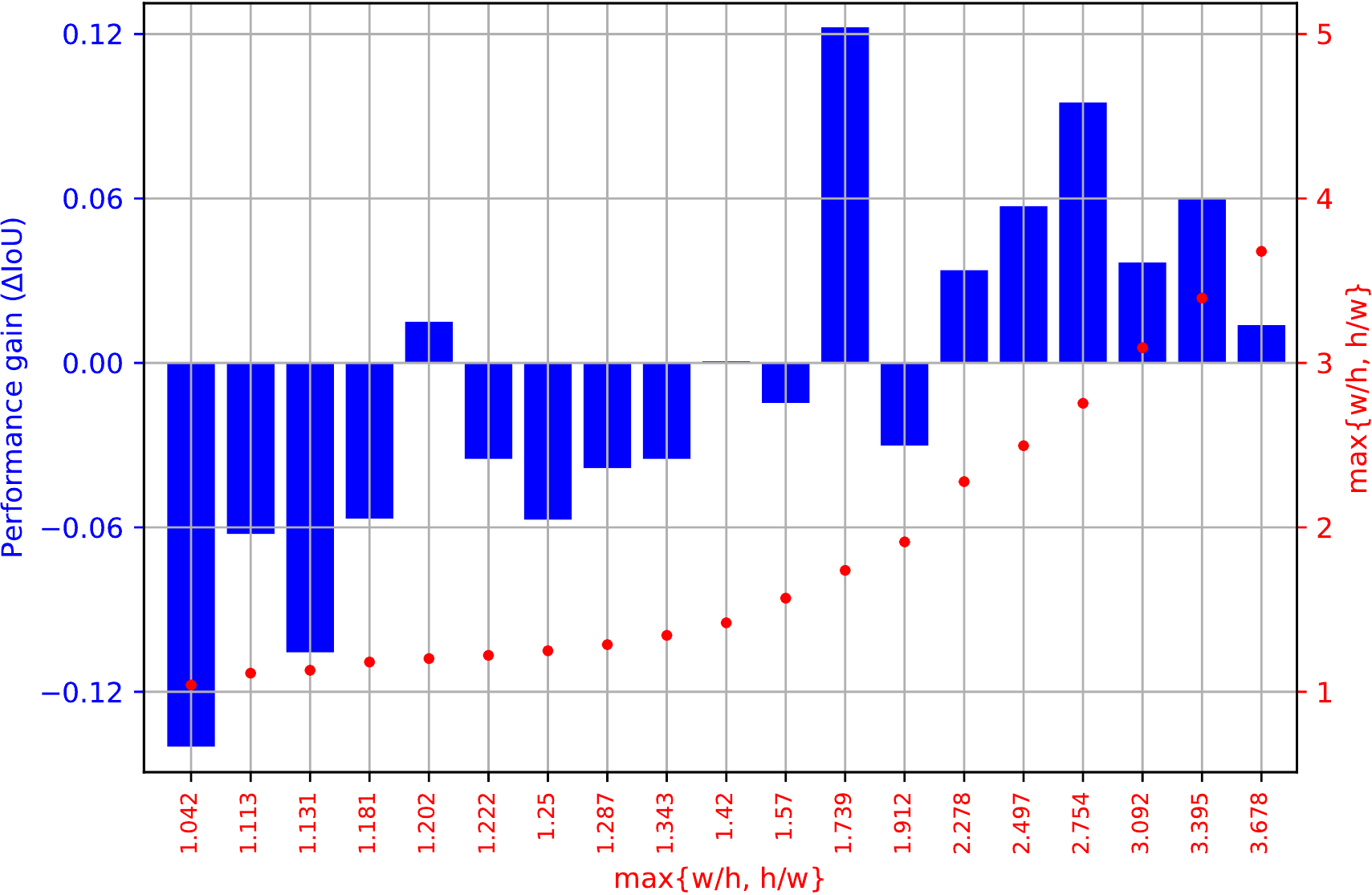}
    \end{center}
    \caption{Performance gain of feature masking is positively correlated with the deviation of aspect ratio from 1.}
\label{fig:mask-perf-gain} 
\end{figure}

Fig. \ref{fig:mask-perf-gain} shows the relationship between the object aspect ratio and the performance gain of spatial masking. Consistent with our observation, when the aspect ratio is far apart from 1, doing spatial masking is helpful. However, when the object is a mediocre one, masking the features is harmful. In general, the performance gain of feature masking is positively correlated with the deviation of aspect ratio from 1.

\setlength{\tabcolsep}{4pt}
\begin{table}
\begin{center}
\caption{Comparing Siam−BM (Mask only) with other high performance and real-time trackers}
\label{table:mask_cmp_stoa}
\begin{tabular}{|c|c|c|c|c|c|}
    \hline
     &\multicolumn{2}{c|}{OTB2013}  &\multicolumn{2}{c|}{OTB100}&\\
    \cline{2-5}
    Trackers & AUC & Prec. & AUC & Prec. & FPS\\
    \hline
    ECOhc \cite{ECO} & 0.652 & 0.874& 0.643& 0.856 & 60\\
    BACF \cite{BACF}  & 0.656 & 0.859& 0.621& 0.822 & 35\\
    PTAV \cite{PTAV} & 0.663 & 0.895& 0.635& 0.849 &25\\
    \hline
    SA-Siam \cite{SASiam} (baseline) & 0.677 & 0.896& 0.657& 0.865&50\\
    Siam-BM (mask only) & \textbf{0.686} & \textbf{0.898}& \textbf{0.662}& \textbf{0.864}&48\\
    \hline
\end{tabular}
\end{center}
\end{table}
\setlength{\tabcolsep}{1.4pt}

\textbf{Siam-BM： }Finally, we show in Tabel~\ref{table:final-combine} how the performance of Siam-BM is gradually improved with our proposed mechanisms. The EAO of the full-fledged Siam-BM reaches 0.335 on VOT2017, which is a huge improvement from 0.287 achieved by SA-Siam. Of course, as we add more mechanisms in Siam-BM, the tracking speed also drops, but the full-fledged Siam-BM still runs in realtime. 

\setlength{\tabcolsep}{4pt}
\begin{table}
\begin{center}
\caption{Analysis of our tracker Siam-BM on the VOT2017. The impact of progressively integrating one contribution at a time is depicted.}
\label{table:final-combine}
\begin{tabular}{|c|ccccccc|}
\hline
& Baseline && Angle && Spatial && Template\\
& SA-Siam &$\Longrightarrow$& Estimation &$\Longrightarrow$& Mask &$\Longrightarrow$& Updating\\
\hline
EAO & 0.287 && 0.301 && 0.322 && \textbf{0.335} \\
Accuracy & 0.529 && 0.544 && 0.551 && \textbf{0.563} \\
Robustness & 1.234 && 1.305 && 1.07 && \textbf{0.977} \\
\hline
FPS & 50 && 35 && 34 && 32\\
\hline
\end{tabular}
\end{center}
\end{table}
\setlength{\tabcolsep}{1.4pt}

\subsection{Comparison with the State-of-the-Art Trackers}
We evaluate our tracker in VOT2017 main challenge and realtime subchallenge. The final model in this paper combines all components mentioned in previous section. We do not evaluate the final model in OTB because the groundtruth labeling in OTB is always upright bounding boxes and applying rotation does not produce a higher IoU even when the tracked bounding box is more precise and tight. 

\begin{figure}[th!]
    \begin{center}
    \includegraphics[width=0.408\columnwidth]{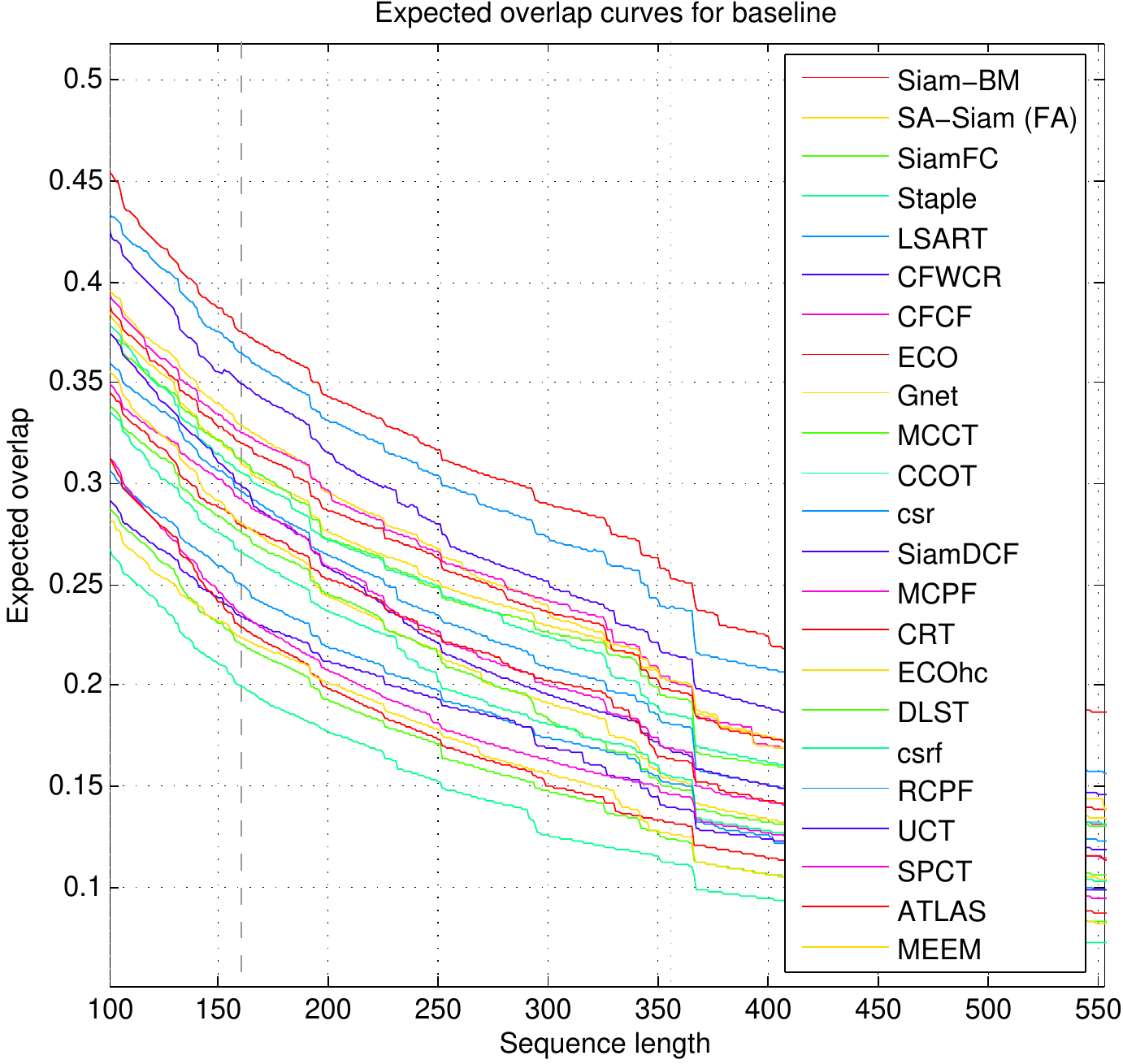}
    \includegraphics[width=0.42\columnwidth]{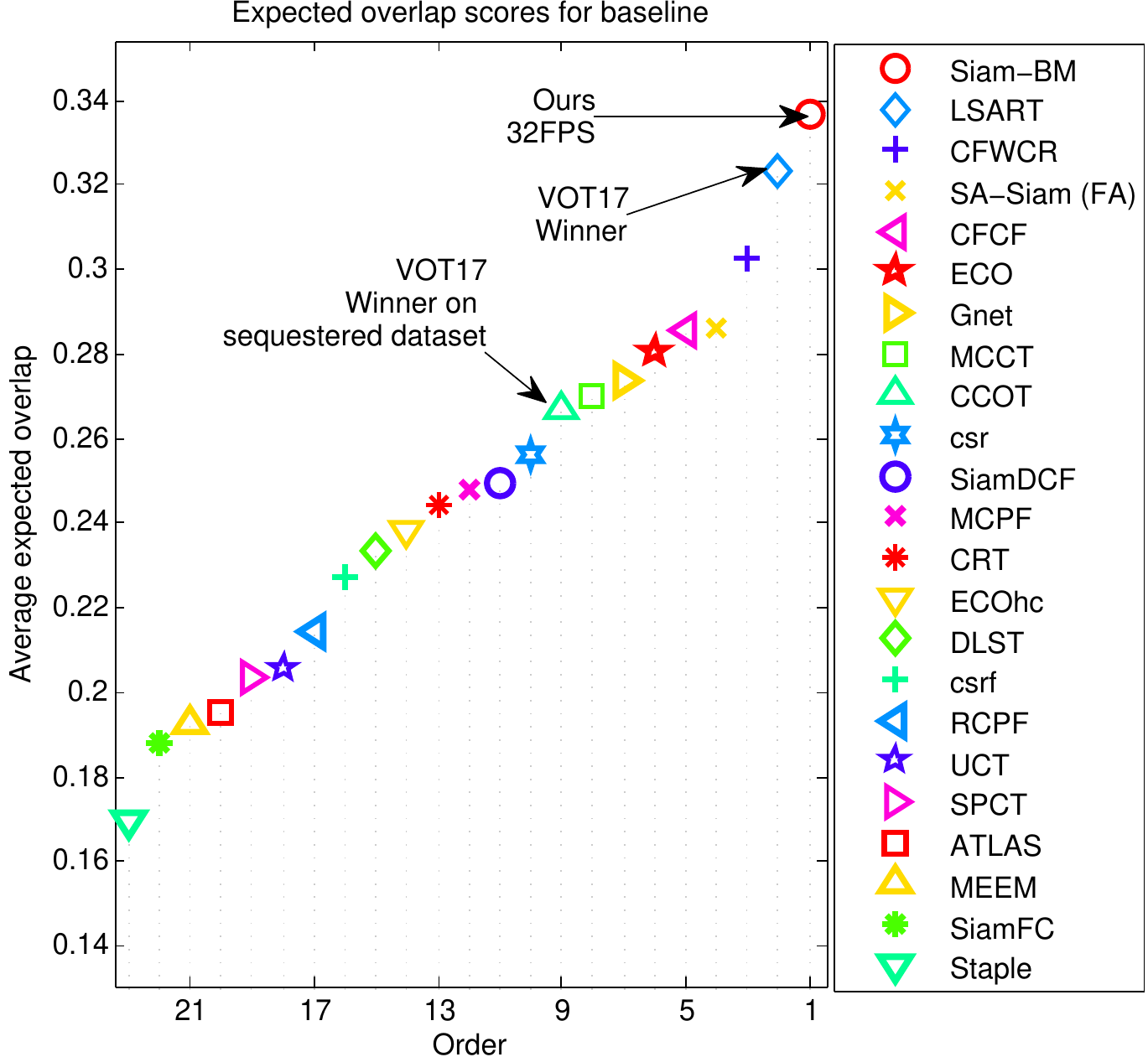}
    \end{center}
    \caption{EAO curves and rank in VOT17 main challenge. FA represents Free Angle here.}
    \label{fig:eao-bl}
\end{figure}

\begin{figure}[th!]
    \begin{center}
    \includegraphics[width=0.4055\columnwidth]{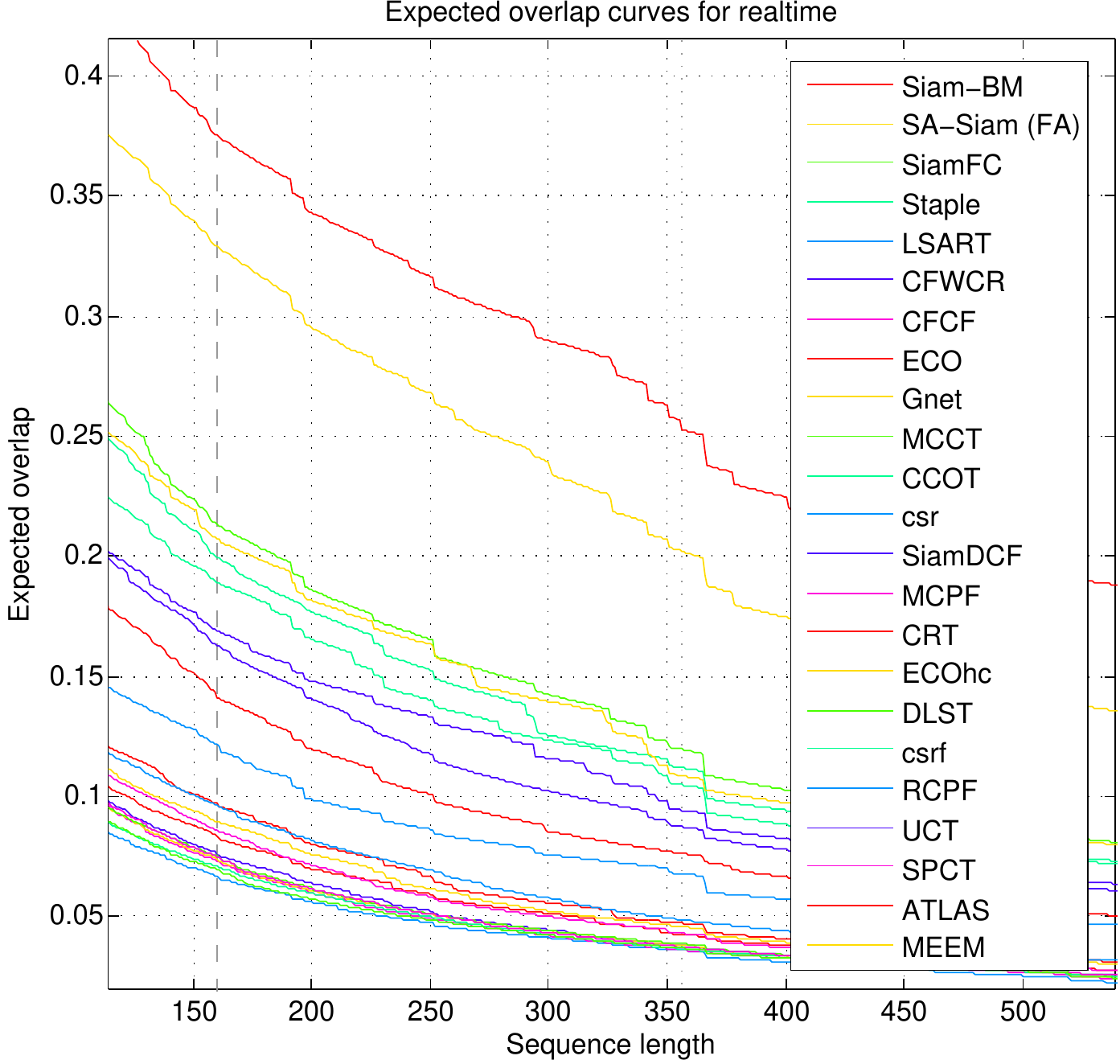}
    \includegraphics[width=0.42\columnwidth]{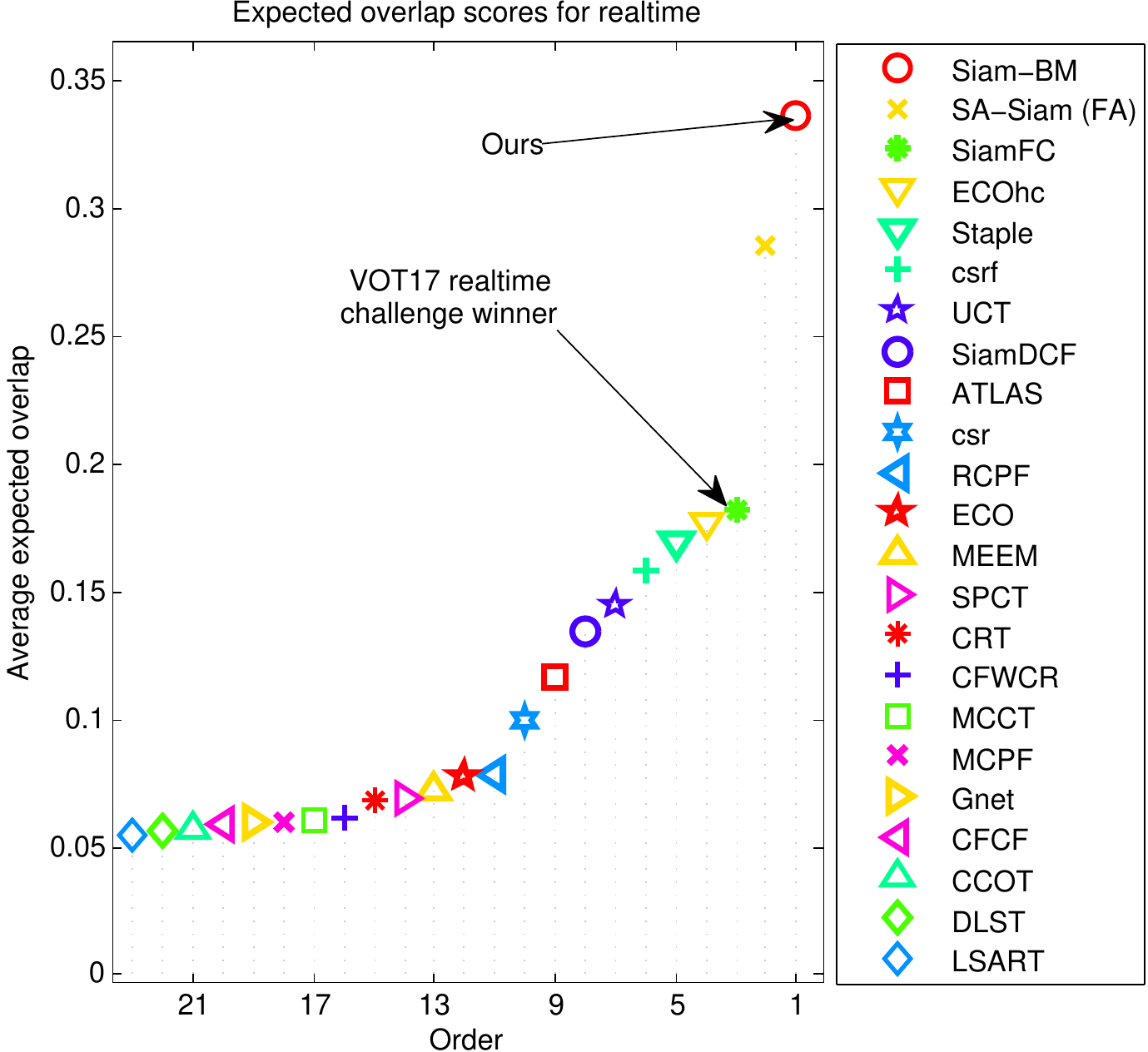}
    \end{center}
    \caption{EAO curves and rank in VOT17 realtime challenge. FA represents Free Angle here.}
    \label{fig:eao-rt}
\end{figure}

\begin{figure}[th!]
    \begin{center}
    \includegraphics[width=0.8\columnwidth]{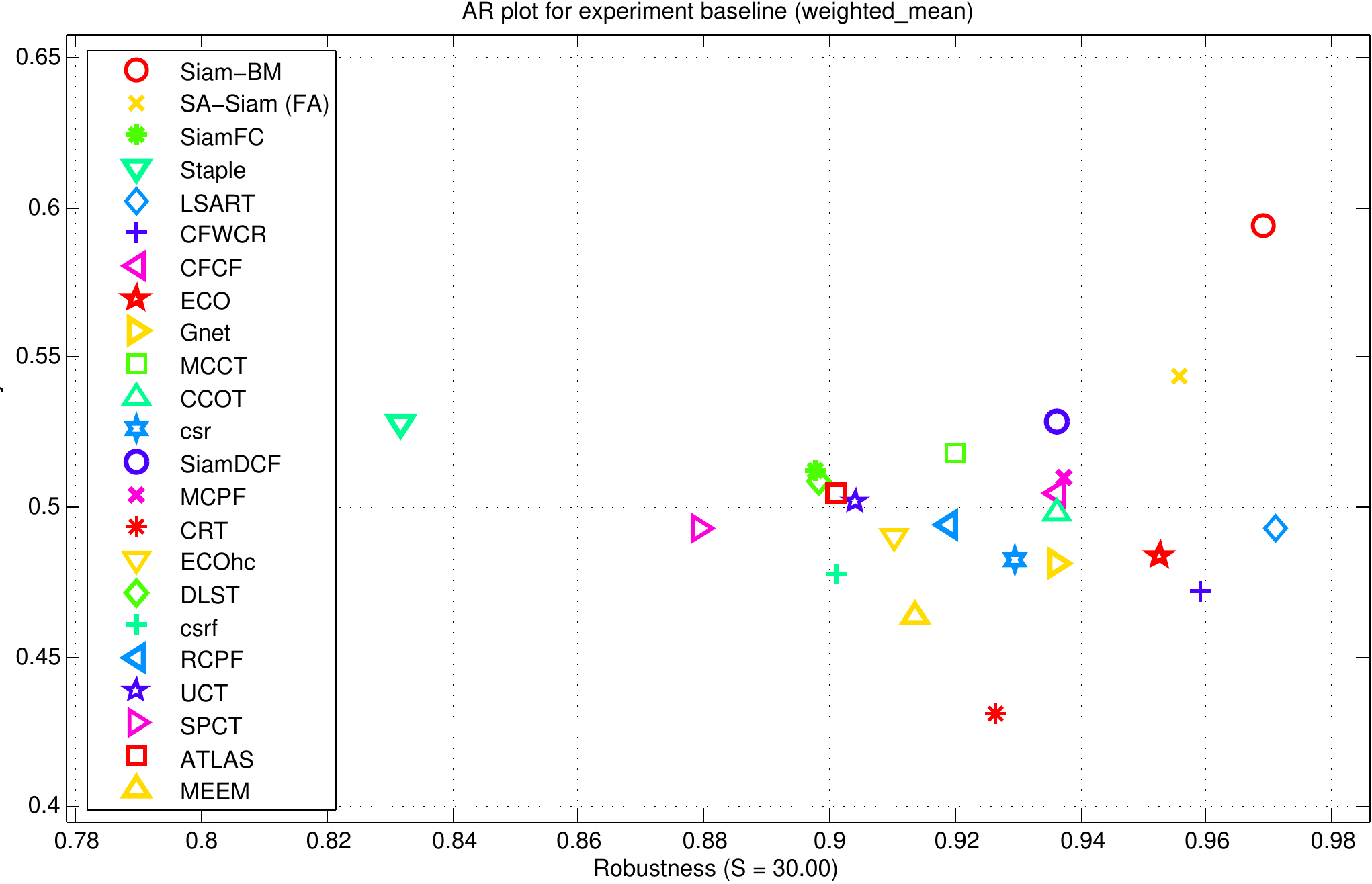}
    \end{center}
    \caption{Accuracy and robustness plots in VOT17 main challenge. Best trackers are closer to the topright corner. FA represents Free Angle here.}
    \label{fig:ar-plot}
\end{figure}

As shown in Fig.\ref{fig:eao-bl}, our Siam-BM tracker is among the best trackers even when non-realtime trackers are considered. From Fig.\ref{fig:eao-rt}, we can see that Siam-BM outperforms all realtime trackers in VOT2017 challenge by a large margin. The Accuracy-Robustness plot in Fig.\ref{fig:ar-plot} also shows the superiority of our tracker. 

We also compare the EAO value of our tracker with some of the latest trackers. RASNet \cite{RASNet} achieves an EAO number of 0.281 in the main challenge and 0.223 in the realtime subchallenge. SiamRPN \cite{SiamRPN} achieves an EAO number of 0.243 in the realtime subchallenge. The EAO number achieved by Siam-BM is much higher. 

\section{Conclusion} \label{sec:conclude}
In this paper, we have designed a SiamFC-based visual object tracker named Siam-BM. The design goal is to achieve a better match between feature maps of the same object from different frames. In order to keep the realtime capability of the tracker, we propose to use low-overhead mechanisms, including parallel scale and angle estimation, fixed spatial mask and moving average template updating. The proposed Siam-BM tracker outperforms state-of-the-art realtime trackers by a large margin on the VOT2017 benchmark. It is even comparable to the best non-realtime trackers. In the future, we will investigate the adaptation of object aspect ratio during tracking.  
\section*{Acknowledgement}

This work was supported in part by National Key Research and Development Program of China 2017YFB1002203, NSFC No.61572451, No.61390514, and No.61632019, Youth Innovation Promotion Association CAS CX2100060016, and Fok Ying Tung Education Foundation WF2100060004.
\clearpage

\bibliographystyle{splncs04}
\bibliography{egbib}
\end{document}